\documentclass[num-refs]{wiley-article}
\usepackage[numbers]{natbib}

\usepackage{caption}
\usepackage{subcaption}
\usepackage{adjustbox}
\usepackage{siunitx}
\usepackage{amsmath}
\usepackage[linesnumbered,ruled,vlined]{algorithm2e}
\let\oldnl\nl
\newcommand{\nonl}{\renewcommand{\nl}{\let\nl\oldnl}}

\papertype{Original Article}
\paperfield{Journal Section}

\title{A Bi-Objective Approach to Last-Mile Delivery Routing Considering Driver Preferences}

\author[1\authfn{1}]{Juan Pablo Mesa}
\author[2\authfn{2}]{Alejandro Montoya}
\author[3\authfn{2}]{Raul Ramos-Pollán}
\author[4\authfn{2}]{Mauricio Toro}

\contrib[\authfn{2}]{Equally contributing authors.}

\affil[1,2,4]{School of Applied Sciences and Engineering, Universidad EAFIT, Medellín, Antioquia, 05001, Colombia}
\affil[3]{Universidad de Antioquia, Medellín, Antioquia, 05001, Colombia}

\corraddress{Juan Pablo Mesa López, School of Applied Sciences and Engineering, Universidad EAFIT, Medellín, Antioquia, 05001, Colombia}
\corremail{jmesalo@eafit.edu.co}

\fundinginfo{}

\runningauthor{Mesa-López et al.}

\begin{document}

\begin{frontmatter}
\maketitle

\begin{abstract}

The Multi-Objective Vehicle Routing Problem (MOVRP) is a complex optimization problem in the transportation and logistics industry. This paper proposes a novel approach to the MOVRP that aims to create routes that consider drivers' and operators' decisions and preferences. We evaluate two approaches to address this objective: visually attractive route planning and data mining of historical driver behavior to plan similar routes. Using a real-world dataset provided by Amazon, we demonstrate that data mining of historical patterns is more effective than visual attractiveness metrics found in the literature. Furthermore, we propose a bi-objective problem to balance the similarity of routes to historical routes and minimize routing costs. We propose a two-stage GRASP algorithm with heuristic box splitting to solve this problem. The proposed algorithm aims to approximate the Pareto front and to present routes that cover a wide range of the objective function space. The results demonstrate that our approach can generate a small number of non-dominated solutions per instance, which can help decision-makers to identify trade-offs between routing costs and drivers' preferences. Our approach has the potential to enhance the last-mile delivery operations of logistics companies by balancing these conflicting objectives.

\end{abstract}
\end{frontmatter}

\section{Introduction}

The Vehicle Routing Problem (VRP) is a well-known problem in the area of combinatorial optimization with applications in many different fields, such as logistics and transportation. It has been studied extensively and has many variations, and extensions \cite{toth2014vehicle}. The VRP is usually formulated as a single-objective problem, where the goal is to route a fleet of vehicles to satisfy the demand of a set of customers while minimizing operations costs. However, in many real-world applications, multiple objectives need to be taken into account. This leads to a variation of the VRP called the multi-objective VRP (MOVRP).

Traditionally, the objective has been to minimize costs. However, in recent years, a new objective in the industry has emerged that focuses on creating routes that take into account the decisions and preferences of operators and drivers.
This objective has been explored from two approaches. Firstly, there has been research on route planning that aims to design visually attractive routes. This line of research is evident in the multiple works \cite{tangandmiller2006, matis2008decision, rossit2019visual, rocha2022visual}. Secondly, another approach involves utilizing data or pattern mining techniques to analyze the historical behavior of drivers and leverage this information for planning similar routes. This approach is evident in the works \cite{krumm2008markov, toledo2013decision, wang2015building, delling2015navigation, canoy2019vehicle, winkenbach2021technical, mesa2023amazon}.

The first approach consists of planning vehicle routes that consider visual attractiveness factors. Even if visual attractiveness is subjective and difficult to express in mathematical terms, in a review by Rossit et al. \cite{rossit2019visual}, the authors identify three features that can be computed, which attractive vehicle routes should have: \textbf{Compactness} or proximity of the customers visited. \textbf{Not overlapping} or crossing of routes with each other. \textbf{Low complexity} of each single route. In this work, we further explore this strategy within a last-mile delivery context.

In 2021, Amazon and MIT's Center for Transportation \& Logistics initiated the Amazon Last-Mile Routing Research Challenge \cite{amazonchallenge}. The goal was to incorporate the know-how of Amazon's experienced drivers into route planning. Participants aimed to predict delivery routes just as these drivers would execute them. According to the results of different participant teams \cite{mesa2021amazon, cook2021amazon, canoy2021amazon}, the objective can be attained by using information extracted from routes previously executed by drivers and integrating this information within traditional routing heuristics. We consider that this approach works well when the executed routes come from planned routes where a route optimization software was involved at some point during the planning. On the contrary, in cases where the real operation routes of a company are planned without any routing optimization step, the route sequences predicted using the previous methods will probably vary considerably with respect to the executed routes.

These objective functions that consider the preferences of drivers and decision-makers can be highly beneficial to them. However, this may conflict with the interests of the company, whose primary concern is minimizing transportation costs. As a result, a bi-objective problem naturally arises, which involves minimizing costs while simultaneously taking driver preferences into account. There is already a study in the literature that addresses this issue, namely the work of Rocha et al. (2022) \cite{rocha2022visual}. In their research, the authors propose the utilization of clustering as an indicator of visual attractiveness in vehicle routing problems. They present a bi-objective vehicle routing problem, examine the impact of visual attractiveness on routing costs, and evaluate the effectiveness of their proposed algorithms in generating VRP solutions that are both cost-effective and visually appealing.

Following the idea of Rocha et al. we propose a bi-objective vehicle routing problem that instead of focusing on the visual attractiveness of the routes, it leverages information from historical routes to plan routes that consider drivers' preferences, while also aiming to minimize routing costs. This work is divided into two parts to demonstrate the potential of our method.

We evaluate and compare both approaches in the first part of this work. We use the model proposed by Mesa et al. \cite{mesa2023amazon} for the approach of leveraging historical data. Additionally, we modify some of the components of this same model to compute visual attractiveness features and use them in the objective function of the other approach. We evaluate both strategies using the dataset and evaluation metrics provided by Amazon \cite{amazon-dataset, amazonchallenge} to determine which method works better for predicting the order in which drivers will execute route sequences. We demonstrate that the method that leverages data from historical routes is more effective than using visual attractiveness metrics.

In the second part of this work, we argue that minimizing routing cost and favoring the routing order of historical routes executions can be conflicting in last-mile delivery. Drivers might be more inclined to execute routes in a familiar or known way. In contrast, the company might prefer that the routes are executed with the minimum possible cost. This leads to a bi-objective problem, with planned routes that are as similar as possible to previous routes and routes that aim to minimize routing costs. Therefore, we formulate the bi-objective problem, and then we propose a simple two-stage GRASP algorithm with heuristic box splitting to solve it. The proposed algorithm aims to approximate the Pareto front \cite{pareto1964} and present routes that cover a wide range of the objective function space to help decision-makers.

The contributions of this work are summarized as follows:

\begin{itemize}
    \item We explore using visual attractiveness metrics to predict route sequences in the way drivers would execute them.
    
    \item We compare the use of visual attractiveness metrics versus data mining strategies for predicting last-mile delivery routes.
    
    \item We propose a relevant last-mile delivery bi-objective problem for decision-makers.

    \item We propose a two-stage GRASP algorithm with heuristic box splitting to approximate the Pareto front of this problem.
    
\end{itemize}


The remainder of this paper is organized as follows. Section \ref{relatedworks} discusses a brief literature review on multi-objective last-mile delivery problems and the prediction of delivery route sequences. Section \ref{motivation} explains the motivation for this research. Section \ref{problemdescription} provides a description of the routing problems that this work studies. Section \ref{solution_method} describes the  three approaches studied in this work and their respective algorithms. Section \ref{experiments} presents the computational experiments and discusses the results. Finally, Section \ref{conclusions} presents the conclusions and future work directions.



\section{Related works}
\label{relatedworks}

This section of the research article will provide an overview of existing methods and studies related to considering decision-makers and drivers' preferences and incorporating cost considerations along with these preferences when planning routes. The review is divided into two parts, each addressing a specific aspect of the problem. The first part explores methods that focus on incorporating drivers' preferences into route planning, while the second part reviews the literature on methods that consider both travel costs and decision-makers' preferences. By reviewing relevant research studies, we aim to identify gaps, challenges, and opportunities for further research in the context of multi-objective last-mile delivery problems and the prediction of delivery route sequences.


In recent years, incorporating the preferences of drivers or decision-makers has gained attention within the vehicle routing community. Several authors have explored an approach that relies on the opinion that practitioners may drive their logistics decisions based on aesthetical considerations \cite{tangandmiller2006, rossit2019visual, rocha2022visual}. Tang and Miller-Hooks \cite{tangandmiller2006} proposed planning visually attractive routes that are compact and do not cross one another. This was achieved by considering some shape measures that would improve the visual layout of routes while maintaining satisfactory results in terms of ``conventional" VRP measures. A comprehensive review of different visual attractiveness metrics used in VRP can be found in Rossit et al. \cite{rossit2019visual}.

Another approach has been recently focused on the last-mile delivery industry, where routes are planned considering data related to the drivers and decision-makers' experience, knowledge, and preferences \cite{toledo2013decision, li2018learning, winkenbach2021technical}. Multiple works have proposed strategies to use data from past routes to learn the routing preferences of drivers \cite{krumm2008markov, wang2015building, delling2015navigation, neto2018combining, canoy2019vehicle}. These works rely on GPS data and Markov models to predict future route decisions. Krumm \cite{krumm2008markov} predicts the next route segment that a driver will take in the current route, while the works of Wang et al. \cite{wang2015building} and Neto et al. \cite{neto2018combining} go a step further and aim to predict the remaining part of a route and the destination.

Similarly, Canoy and Guns \cite{canoy2019vehicle} also used weighted Markov models to plan multiple routes in VRP considering driver's preferences, while avoiding the need to specify drivers' preferences explicitly. Additionally, Canoy et al. \cite{canoy2021amazon} extended the use of Markov models and transition probabilities together with distance information to solve the Traveling Salesman Problem (TSP) in the Amazon Routing Challenge \cite{amazonchallenge}. The approach of Canoy et al. focuses on vehicle route planning with constraints and how to learn the implicit constraints that can be extracted from past routes. Other solutions that were presented in the Amazon Challenge also leveraged the use of constraints extracted from past routes. Some notable ones were the solutions of the three winning teams of Cook et al. \cite{cook2021amazon}, Guo et al. \cite{guo2021amazon}, and Arslan and Abay \cite{arslan2021amazon}. All these solutions encoded patterns found in the data either as clustering constraints, precedence, or antecedence constraints and coupled them with powerful local search heuristics. Among the other competitors, Mesa et al. \cite{mesa2021amazon} also used the strategy of constructing a probability transition matrix based on historical trips taken by a driver to predict a route. However, instead of using the extracted information as constraints for route planning, the information was considered within the objective function as penalizations. Instead of relying on a Markov model to calculate probabilities, the authors used vector fields to represent historical data. They used the directions and magnitudes of the vectors to calculate the transition probabilities. Additionally, they developed a two-stage approach for the TSP where one first calculates the transition probabilities between clusters based on historical data and then uses these probabilities, along with other factors, within a TSP metaheuristic to predict route sequences. This approach was later improved by the same authors in \cite{mesa2023amazon}. 

The problem becomes more complex when multiple objectives need to be considered, such as minimizing transportation costs, reducing vehicle emissions, improving customer satisfaction, or including the drivers' preferences. These objectives may conflict with each other, and finding an optimal solution that satisfies all of them is challenging \cite{jozefowiez2008multi}. Multi-objective optimization techniques aim to find multiple solutions, called Pareto optimal solutions, that represent trade-offs between conflicting objectives. This approach provides decision-makers with a range of alternatives from which they can choose the most suitable solution based on their preferences.


Some companies aim to optimize vehicle routes to reduce costs, while also considering consistency in visiting each customer at similar times of the day with familiar drivers to increase customer satisfaction and lifetime value. Lian et al. \cite{lian2016improved} proposed a multi-objective strategy to reduce the maximum number of drivers visiting a customer to improve driver consistency and minimize the maximum difference between the earliest and latest arrival time to achieve arrival time consistency. The authors included consistency considerations such as driver and time consistency as objectives in their multi-objective variant of the Consistent Vehicle Routing Problem. By including these considerations as objectives rather than constraints, their approach can help decision-makers balance competing objectives effectively, allowing them to choose between a minimal-cost routing plan with poor consistency, a high-cost plan with perfect consistency, or any plan in between.

Rocha et al. \cite{rocha2022visual} explore delivery cases where incorporating visual attractiveness into route planning can lead to more favorable routes for drivers and present a bi-objective vehicle routing model to support their proposal. They proposed the use of customer clustering as a proxy for visual attractiveness in vehicle routing problems, assessing the impact of visual attractiveness on routing cost and evaluating the effectiveness of NSGA-II heuristics in producing low-cost and visually attractive VRP solutions.


In conclusion, there are two research gaps that this work aims to address. Firstly, while previous works have separately explored methods based on data from driver experience or visual attractiveness metrics for route planning, this paper stands out as the only one that directly compares these two approaches using routes from a real-life last-mile delivery operation. By conducting this comparison, we can determine which method yields better results in terms of route prediction based on driver preferences. Secondly, this work introduces a novel perspective by formulating the problem as a bi-objective optimization task that considers both the total cost of the route and the objective of predicting the route based on historical data and machine learning models. To the best of our knowledge, this is the first study that tackles a bi-objective problem in this context, offering decision-makers valuable insights into optimizing their route planning process.

\section{Motivation}
\label{motivation}

In 2021, Amazon and MIT's Center for Transportation \& Logistics launched the 2021 Amazon Last-Mile Routing Research Challenge \cite{amazonchallenge}. The challenge focused on improving last-mile delivery by addressing the issue of drivers deviating from planned routes. Traditionally, route planning does not consider driver preferences and their tacit knowledge of the areas they traverse. To bridge this gap, the challenge encouraged the use of machine learning to devise or predict route sequences that mirror the choices of seasoned drivers.

The routes in the dataset provided in the challenge are characterized for being a TSP solution that ends at the same station from which it departed, with asymmetric travel times, given that the TSP instances come from real-world routing operations, and each stop has an assigned zone ID. In previous studies from \cite{arslan2021amazon, cook2022constrained, mesa2023amazon}, different authors determined that the drivers usually respect the zone ordering and visit the stops with the same zone ID contiguously before visiting other stops with a different zone ID. This routing with zones is considered in the literature as a Clustered TSP. Therefore, in this work, we are going to address the routing problem as a Clustered Asymmetric TSP.

\begin{figure}
     \centering
     \begin{subfigure}[b]{0.3\textwidth}
         \centering
         \includegraphics[width=\textwidth]{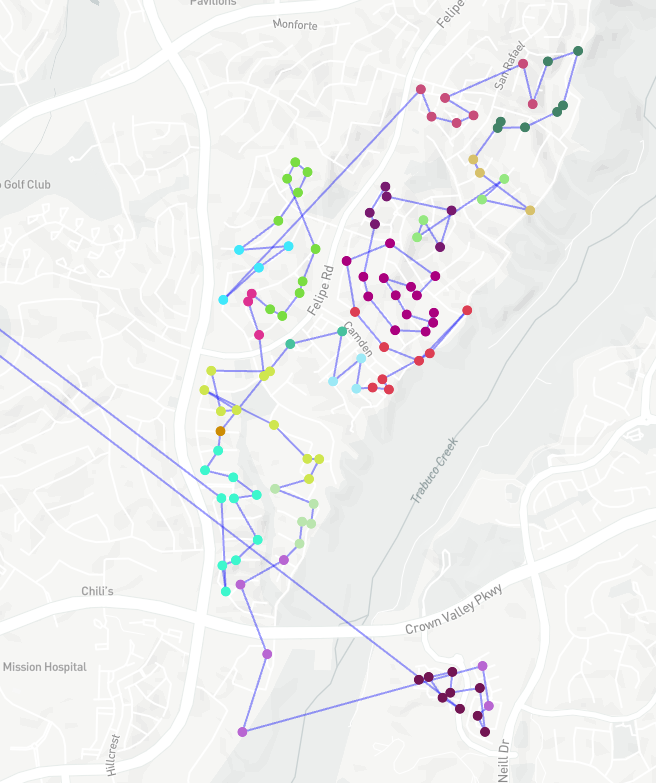}
         \caption{Planned Clustered ATSP route with minimized traveling time of 172 minutes.}
         \label{fig:Route_9_planned_by_us}
     \end{subfigure}
     \hfill
     \begin{subfigure}[b]{0.3\textwidth}
         \centering
         \includegraphics[width=\textwidth]{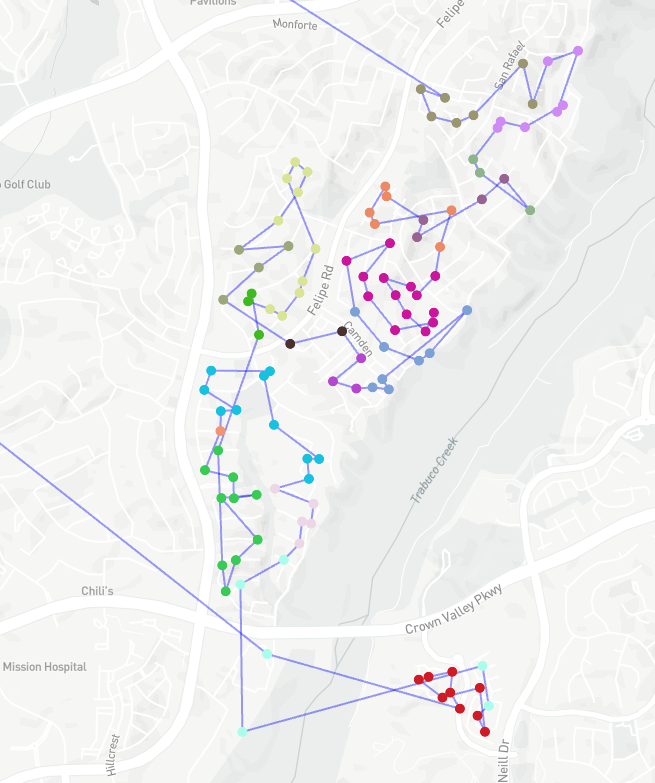}
         \caption{Clustered ATSP route as planned by Amazon with traveling time of 177 minutes.}
         \label{fig:Route_9_planned_by_amazon}
     \end{subfigure}
     \hfill
     \begin{subfigure}[b]{0.3\textwidth}
         \centering
         \includegraphics[width=\textwidth]{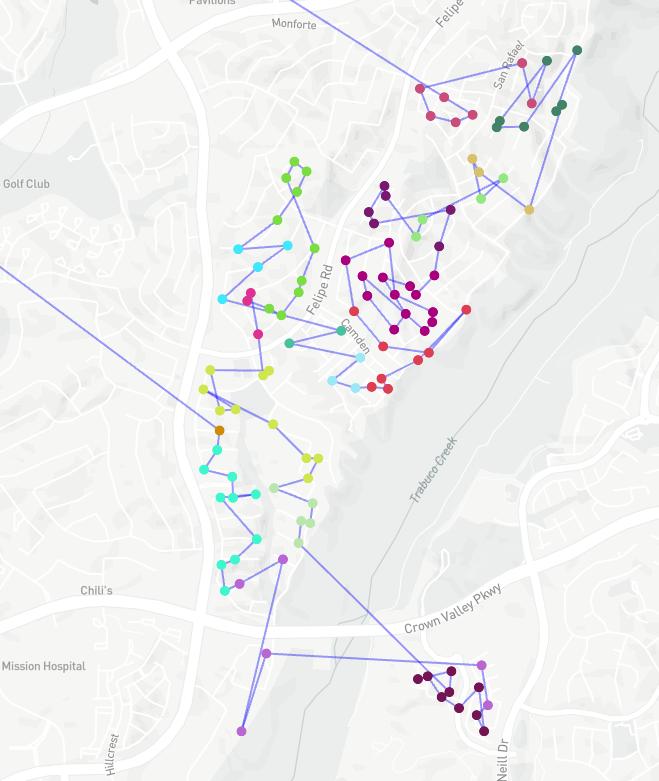}
         \caption{Planned Clustered ATSP route similar to Amazon's route with traveling time of 174.5 minutes.}
         \label{fig:Route_9_planned_by_intermediate}
     \end{subfigure}
        \caption{Different routing solutions for the same instance of the Amazon dataset.}
        \label{fig:three amazon plots}
\end{figure}

The research challenge pushed participants to propose techniques that aimed to capture the decision-making and tacit knowledge of seasoned delivery drivers and generate route sequences that a driver would consider convenient, efficient, and safe. However, we consider that even if the challenge affronted a reality in last-mile delivery routing, it still lacked considering the minimization of the total cost of the route, which is one of the main objectives that the industry considers. For example, using data from the Amazon Last-Mile Routing Research Challenge and a VRP solver, we obtain the routes observed in Figure \ref{fig:three amazon plots}.

In Figure \ref{fig:Route_9_planned_by_us}, we show the solution with the minimal traveling time cost that the solver was able to find for this problem, with a traveling time of 172 minutes. A colored circle represents each stop, and each color represents a different zone ID. However, an Amazon driver performed the route as it is shown in Figure \ref{fig:Route_9_planned_by_amazon}. The traveling time cost of this route is 177 minutes. The drivers or the route planners might give different arguments, such as this route is safer or easier to execute, which can justify the extra travel-time cost. A third possible route is shown in Figure \ref{fig:Route_9_planned_by_intermediate}, with a traveling time cost of 174.5 minutes, which is longer than the first route, but shorter than the way in which an Amazon driver performed the route. One can argue that this route is more similar to the Amazon route than the route in Figure \ref{fig:Route_9_planned_by_us} in terms of a similarity score metric that will be explained later in this work.

It can be observed that predicting or planning the routes that consider the preferences of the drivers does not necessarily translate into minimizing the cost of the routes. Therefore, the problem that we aim to solve should consider both objectives simultaneously. However, it is important to determine how to measure the amount of ``preference" that a driver has toward a route. In the literature, we found two approaches. One approach is the use of visual attractiveness metrics \cite{rossit2019visual} to evaluate and create routes based on the ``attractiveness'' of the routes to drivers and decision-makers. The other approach consists of using data mining strategies to extract historical patterns about the order in which stops were visited along the routes; the patterns are then used as constraints or penalizations to guide the route planning.

\section{Problem description}
\label{problemdescription}

The Clustered Asymmetric Traveling Salesman Problem Problem (CATSP) can be formulated using graph notation as follows. Let's consider a directed graph $G = (V, A)$, where $V$ is the set of vertices representing stops and $A$ is the set of arcs representing directed edges between stops. There is one depot station $s$, from which a route must start and end, $n$ stops, and $m$ clusters. 

Let $Z$ be the set of clusters, and $V_g$ be the set of vertices in cluster $g$ for $g = 1, 2, \ldots, m$, such that $\bigcup_{k=1}^{m} V_g = V$ and $V_i \cap V_j = \emptyset$ for $i \neq j$. Let $a_{ij}$ be the arc from vertex $i$ to vertex $j$ for $i, j \in V$ and $i \neq j$. The arc set $A$ is defined as $A = \{(i, j): i, j \in V, i \neq j\}$, and the cost of traversing an arc $(i, j)$ is $c_{ij}$. Let $x_{ij}$ be a binary decision variable indicating whether arc $(i, j)$ is included in the tour. $x_{ij} = 1$ if arc $(i, j)$ is selected, and $x_{ij} = 0$ otherwise.

The ``traditional" objective of the CATSP is to minimize the total cost of the tour, which can be represented as:

\begin{equation} \label{eq:general_objective}
    \min f_o = \sum_{i \in V}\sum_{j \in V}\, c_{ij}x_{ij}
\end{equation}

Subject to the following constraints:

1. Each stop should be visited exactly once:
\[
\sum_{j \in V, j \neq i} x_{ij} = 1 \quad \forall i \in V
\]

2. Each stop should be left exactly once:
\[
\sum_{i \in V, i \neq j} x_{ij} = 1 \quad \forall j \in V
\]

3. Elimination of subtours:
\[
\sum_{i \in S}\sum_{j \in S, j \neq i} x_{ij} \leq |S| - 1 \quad \forall S \subset V, |S| \geq 2
\]

4. Stops within each cluster should be visited consecutively:
\[
\sum_{i \in V_g}\sum_{j \in V_g} x_{ij} = |V_g|-1 \quad \forall V_g \subset V, |V_g| \geq 1, g = 1, 2, \dots, m
\]

5. Binary variable constraints:
\[
x_{ij} \in \{0, 1\} \quad \forall i \in V, \forall j \in V, i \neq j
\]

The objective is to find the values of \(x_{ij}\) that minimize the objective function while satisfying the given constraints.

The 2021 Amazon Last Mile Routing Research Challenge provided a dataset of real-life operations with over 6000 last-mile delivery route instances. Each instance consists of a collection of \textit{stops} that must be visited by a single driver. Additionally, each instance contains information about a vehicle driver's sequence of stops. Each stop has attributes such as latitude and longitude, delivery time window, and zone ID (cluster to which the stop should belong). The asymmetric travel times between stops in a route are also provided on the research challenge dataset introduced by Merchan et al. \cite{amazon-dataset}.

In this work, for each instance, we are going to aim to minimize the total travel time of the route as a first objective. As the second objective, we aim to predict routes that a driver would prefer to execute. For this second objective, we are going to evaluate the two approaches that were previously mentioned. One approach is the use of visual attractiveness metrics \cite{rossit2019visual} to evaluate and create routes based on the ``attractiveness" of the routes to drivers and decision-makers. The other approach consists in using data mining strategies to extract historical patterns about the order in which stops were visited along the routes. The patterns are then used as constraints or penalizations to guide the route planning. Later in this work, we will demonstrate with computational experiments which of these two approaches works better for predicting routes in the order that drivers would prefer to execute them. In some cases, finding vehicle routes that optimize both objectives simultaneously will prove to be impossible. Therefore, the problem we tackle is to approximate the \textbf{Pareto front} \cite{pareto1964} for each instance.

\section{Solution method}
\label{solution_method}

In this section, we propose a solution method for the problem involving the CATSP and the previously mentioned objectives. The solution method is subdivided into four sections, each addressing a specific aspect of the problem. Firstly, we present a two-stage algorithm as a general framework for solving the CATSP, which aims to minimize the total travel time. This algorithm serves as the foundation for subsequent adaptations. The following subsection explains how the method is adapted to incorporate visual attractiveness metrics. We discuss the specific visual attractiveness metrics used and the composition of the objective function. Additionally, we explore the adaptation of the method to incorporate patterns found in historical routes based on ML methods. Finally, we address the bi-objective nature of the problem. Regardless of whether the second objective is based on visual attractiveness or ML-based, we aim to find solutions that conform to the Pareto frontier. Thus, we showcase how the general framework is modified to handle the bi-objective problem effectively.

\subsection{Two-stage GRASP algorithm for the CATSP}

We start by introducing a multi-start-based metaheuristic as a solver for the CATSP with the objective of minimizing total route time. The solver is a Two-Stage Greedy Randomized Adaptive Search Procedure (GRASP) based on the method introduced by Mesa et al. \cite{mesa2023amazon}.

The CATSP solver is divided into two main steps to generate route sequences: ordering the clusters and ordering the stops within the cluster. First, we compute the average travel times between clusters (zones) and use these times to solve a TSP of clusters. Then, after the order of the clusters has been fixed, the stops are randomly assigned to each cluster, and a TSP of all the stops is solved with heuristics that only allow intra-clusters movements. The pseudo-code of the CATSP solver is presented in Algorithm \ref{algo:general_catsp_solver}.

\begin{algorithm} [h]
 
	\KwIn{$V, T, L, f_o$}

    $cost_{best} \leftarrow \infty$
  
	$V$ $\leftarrow$ Assign missing zone IDs to stops ($V$)

    $M = \emptyset \leftarrow $Matrix of average cluster-to-cluster (zone-to-zone) travel times
    
    \For{$g = 1, 2, \dots, m$}
	{
 
        \For{$h = 1, 2, \dots, m \quad h \neq g$}
	   {

            \For{$i \in V_g$}
	       {
 
             \For{$j \in V_h$}
	           {

                    $M_{gh} = M_{gh} + T_{ij}$
    
                    }

                }
            $M_{gh} = \frac{M_{gh}}{|V_g| + |V_h|} $
    
            }

    }

	\For{$iteration \leq L$}
	{   
	    Zones solution $z \leftarrow$ Random\_order($Z$)
		
		$z \leftarrow$ relocate\_zones($z, M$)
		
		Solution $x \leftarrow$ Randomize ($z, N$)
		
		$x \leftarrow$ VND($x, neighborhoods, T$)
		
		\If{$f_o(x) < cost_{best} $}
		{
			$cost_{best} \leftarrow f_o(x)$
			
			$x_{best} \leftarrow x$
		}

	}
 
	\Return{$x_{best}$}
	
	\caption{CATSP Solver. Adapted from \cite{mesa2023amazon}}
    \label{algo:general_catsp_solver}
\end{algorithm}

Algorithm \ref{algo:general_catsp_solver} describes the proposed Two-Stage GRASP. ``The input parameters $V$ and $T$ correspond to the set of stops and the travel times between stops, respectively. The input parameter $L$ indicates the maximum number of iterations performed by the GRASP. Line 1 initializes the cost of the best incumbent solution to infinite. Then in Line 2, when we \textbf{Assign missing zone IDs to stops}, to each stop that does not have a \textit{zone ID}, we assign it the zone ID of the nearest stop (that does have a zone ID) in terms of travel time. After that, in Lines 3-9, we use the travel time information between the stops to calculate the estimated travel time $M_{gh}$ between each cluster $g$ to cluster $h$, which we call zone-to-zone (or cluster-to-cluster) travel time. Lines 10-17 execute the main loop of the GRASP for $L$ iterations. Line 11 of the algorithm creates an initial zones solution route $z$, composed of each cluster (zone) and the depot station $s$. The depot is assigned at the beginning and the end of route $z$, while all the clusters are randomly ordered. Next, line 12 applies the classical relocate neighborhood with a first-improvement strategy, using the cluster-to-cluster travel time information from $M$ to calculate the cost of the solution. When no further relocate movements that would improve the cost of $z$ can be applied, the algorithm proceeds to line 13. The algorithm constructs an initial solution $x$ of stops $i \in V$, assigning all stops to their corresponding cluster in a randomized order while keeping the order of the clusters from solution $z$. After constructing the initial solution $x$, the algorithm applies the local search phase with a Variable Neighborhood Descent (VND) \citep{mladenovic1997variable} in line 14. In lines 15 to 17, the overall best solution is updated when a better solution (lower objective function value) is found. Finally, line 18 returns the best solution after all iterations are performed. The VND in this GRASP uses the following neighborhoods with a best-improvement configuration. \textit{Relocate stop} relocates the position of one stop at a time within the same zone. \textit{Swap stops} exchanges the position of two stops from the same zone."\cite{mesa2023amazon}

\subsection{Adapting the CATSP solver with visual attractiveness metrics}
\label{sec:adapting_solver_visual_metrics}

We can modify the CATSP solver, whose objective function is the minimization of total travel time, to consider visual attractiveness metrics in an attempt to plan routes preferred by drivers or decision-makers. In this section, we explain the visual attractiveness metrics used, the changes to the objective function, and how the CATSP solver is adapted to incorporate these metrics.

The visual attractiveness metrics are usually formulated in the literature for VRP variants. Rossit et al. \cite{rossit2019visual} identify three common characteristics that many authors consider that attractive vehicle routes should have: \textbf{compactness, non-overlap, and no-complexity}. Different authors have proposed multiple metrics for each characteristic. However, given that in this work we are tackling a CATSP, we selected four metrics that can be applied to individual routes. The visual attractiveness metrics $\upsilon_k$ that we use are as follows.

\begin{itemize}
    \item \textbf{Average Distance Compactness} introduced by Matis \cite{matis2008decision}: 
    \begin{equation}
        \upsilon_{adc} = \frac{AvgDist}{AvgMaxDist}
    \end{equation}
     $AvgDist$ is the average distance (or travel time) between two consecutive stops in the route, and $AvgMaxDist$ is the average distance of the 20\% longest distances between two consecutive stops in the route. A larger $\upsilon_{adc}$ indicates that the route is more compact.

    \item \textbf{Center Distance Compactness} introduced by Kant et al. \cite{kant2008coca} :
    \begin{equation}
        \upsilon_{cdc} = \sum_{i \in V}{distance(i, c)}
    \end{equation}
    $c$ is the stop located at the center (intermediate position) of the route. This work represents $distance()$ by travel time between stops. A smaller $\upsilon_{cdc}$ indicates that the route is more compact.

    \item \textbf{Number of crossings} introduced by Poot et al. \cite{poot2002savings}:

    Indicates the total number of crossings ($\upsilon_{nc}$) between arcs of the route. This value is an approximation of the real number of crossings, as the coordinates of the stops are used to determine the crossings geometrically instead of using a real network of roads. A smaller number of crossings indicates more visual attractiveness.

    \item \textbf{Bending energy} introduced by Gretton and Kilby \cite{gretton2013study}:
    \begin{equation}
        \upsilon_{be} = \frac{\sum_{i = 2}^{|V|}{(alpha_{i-2, i-1, i})}}{|V|}
    \end{equation}
    We take the vectors formed by the stops $i-2$, $i-1$, and $i$, and compute the smallest angle $alpha_{i-2, i-1, i}$, in radians, between these vectors. The smaller the sum of these angles, the more visually attractive.
    
\end{itemize}

Each one of these visual attractiveness metrics $\upsilon_k$ can be considered within the objective function when solving the CATSP. We can modify the objective function from Equation \ref{eq:general_objective} by adding the metric that we want to consider, as indicated in Equation \ref{eq:visual_attractiveness_objective}. If the metric that we want to consider is the Average Distance Compactness $\upsilon_{adc}$, we should subtract it instead of adding it because, contrary to the other metrics that we want to minimize, that metric indicates more visual attractiveness when it is maximized.

\begin{equation} \label{eq:visual_attractiveness_objective}
    \min  f_o = \upsilon_k + \sum_{i \in V}\sum_{j \in V}\, c_{ij}x_{ij}
\end{equation}

With the new formulation of the objective function and the means to calculate the visual attractiveness of a route, we can modify Algorithm \ref{algo:general_catsp_solver}. We insert the visual attractiveness metric $\upsilon_k$ computation immediately after Line 14. Additionally, we modify lines 15 and 16 to include $\upsilon_k$ within the computation of the objective function $f_o(x)$, which now is $f_o(x, \upsilon_k)$. The pseudo-code of the CATSP solver with visual attractiveness metrics is presented in the Appendix in Algorithm \ref{algo:catsp_solver_visual_metrics}.

\subsection{Adapting the CATSP solver with data mining from historical routes}
\label{sec:adapting_solver_historical_routes}

Another way in which we can modify the CATSP solver to plan routes preferred by drivers or decision-makers is to consider information that can be extracted from historical routes. This section explains the method for extracting historical patterns from routes, the changes applied to the objective function, and how the CATSP solver is adapted to incorporate these metrics.

There are multiple approaches that use data mining techniques to find patterns in delivery routes \cite{krumm2008markov, delling2015navigation, canoy2019vehicle, winkenbach2021technical, mesa2023amazon}. For this work, we selected the strategy proposed by Mesa et al. \cite{mesa2023amazon}. The strategy consists in extracting information from historic routes using a probability estimation method that learns transition probabilities between clusters and then incorporates these transition probabilities into route planning algorithms as penalizations on the objective function. This strategy of data mining (DM) information from routes works as follows.

We first define a \textit{zone centroid} for each cluster/zone by taking the average latitude and longitude coordinates of the stops within that zone. A vector between each zone centroid and the next zone centroid in a route can be traced to represent the route. If we do this for every route in a city, we will construct a vector field across the city using historical routes. However, each of these vectors might have different magnitudes. Therefore, we divide them into \textit{step vectors}, each with a length determined by the parameter $\beta$. This field of step vectors enables us to estimate a preferred heading for any arbitrary point on the 2D plane by averaging the observed directions of vectors in its vicinity.

We then construct a vector field that comprises all step vectors from the observed sequences of zones in the available historical routes. Let $Q$ be the set of historic routes. From $Q$, we derive the set of step vectors $V_j$ associated with the sequence of zones in each route $j \in Q$. Consequently, we define the vector field $F$ as the union of all $V_j$ sets, i.e., $F = \bigcup_{j \in Q} V_j$. Each step vector $\mathbf{v}$ in $F$ is characterized by an origin $o_v$, a final position $f_v$, and a length of $\beta$ meters. 

Once we have a vector field constructed with the data from all available historical routes, we can use the information to construct matrices of transition probability matrices and deviation probability for each new route that we want to plan. For a given route $x$ to be planned, a transition probability matrix $H_{ij}$ represents the probability that zone $i \in Z$ is visited before zone $j \in Z$. The matrix of deviation probability $A_{ij}$ represents the probability that the direction taken by heading from zone $i$ to zone $j$ deviates from the historical direction taken after visiting zone $i$. These matrices are calculated as follows.

First, we introduce the concept of \textit{heading vectors}. These vectors point in the direction where the route sequence should go for each zone centroid of zone $p \in Z$ after visiting all the stops in the current zone. The estimation of a heading vector $\mathbf{h}(p)$ for zone $p$ is obtained using Equation (\ref{eq:heading_estimation}), where $\alpha$ is a weighting factor parameter and $D(p, o_v)$ represents the distance from zone centroid $p$ to the origin $o_v$ of a step vector $\mathbf{v}$.

\begin{equation} \label{eq:heading_estimation}
    \mathbf{h}(p) = \sum\limits_{\mathbf{v} \in F} \mathbf{v} \cdot e^{-D(p, o_v)\cdot \alpha}
\end{equation}

To efficiently compute the heading vectors, we take advantage of the fact that the contribution of a step vector $\mathbf{v}$ diminishes as it moves further away from the zone centroid $p$. This allows us to focus on step vectors located within a maximum radius Euclidean distance from $p$. We employ the KDTree algorithm \citep{kdtree1975}, based on a binary tree data structure, to store the origins of the step vectors. By dividing the vector space in half at each tree level, the algorithm enables us to find all step vectors $\mathbf{v}$ within the maximum distance $max_d$ (determined such that $ e^{max_d}<10^{-3}$) from each zone centroid $p \in Z$ efficiently using a single query.

Once we have computed the heading vectors of a new route, we can estimate the transition probability matrix $H_{ij}$ using Equation (\ref{eq:heading_prob}). ``This equation takes into account the normalized Euclidean distance $\delta_{ij}$ between zone centroids $i$ and $j$, as well as the normalized cosine distance $\epsilon_{ij}$ between the estimated heading vector $\mathbf{h}(i)$ at zone $i$ and the direction vector from zone $i$ to zone $j$. Both distances are normalized based on the maximum distances between any pair of zone centroids within the route. It is important to note that when there is no historical information available regarding the order between zones $i$ and $j$, we assign a value of $0.5$ to $H_{ij}$ to avoid any bias in their sequencing."\cite{mesa2023amazon}

\begin{equation} \label{eq:heading_prob}
    H_{ij} = \frac{1}{2}\delta_{ij} + (1-\delta_{ij})\epsilon_{ij}
\end{equation}

Using the heading vectors, we can also estimate the deviation probability matrix $A_{ij}$. First, we estimate a direction vector $u_{ij}$ by subtracting the zone centroid of $i \in Z$ from the zone centroid of $j \in Z$ for each zone in $Z$. Using the direction vector, we can calculate the deviation angle $a_{ij}$ (in degrees) between $u_{ij}$ and the heading vector originating from the zone centroid of $i$. This deviation angle is used in Equation (\ref{eq:deviation_prob}) to construct the deviation probability matrix. Notably, when $A_{ij} = 1$, this indicates that the direction the route takes after visiting zone $i$ is probably deviating from the historical route order. Conversely, when $A_{ij} = 0$, the route follows the estimated historical direction.

\begin{equation} \label{eq:deviation_prob}
    A_{ij}=\frac{a_{ij}}{180}
\end{equation}

We can also leverage data from the zone centroids that we construct for the new route. We can estimate the travel time $r_{ij}$ between the zone centroids of $i \in Z$ and $j \in Z$ using Equation (\ref{eq:Oij}). In this equation, $N_i$ and $N_j$ represent the sets of stops associated with zone $i$ and $j$ respectively, and $t_{kl}$ denotes the travel time between stop $k \in N_i$ and stop $l \in N_j$. The zone-to-zone travel time $M_{ij}$ is calculated as the sum of all travel times between stops in $N_i$ and $N_j$, divided by the total number of stops in both sets. This calculation can easily be extended to include the estimated travel time from each zone to the delivery station $s$, assuming $s$ to represent a zone with only one stop.

\begin{equation} \label{eq:Oij}
    M_{ij} = \frac{\sum\limits_{k \in N_i}\sum\limits_{l \in N_j} t_{kl}}{|N_i|+|N_j|}
\end{equation}

We now have the matrices $M_{ij}$, $H_{ij}$, and $A_{ij}$ that contain information about route $x$ and from the historical routes that were executed nearby route $x$. We incorporate this information to plan a route as follows. 

Using the route $x$, we can estimate the binary variable $m_{ij}$, which is equal to 1 if zone $i \in Z$ immediately precedes zone $j \in Z$, and 0 otherwise. We can estimate the deviation from the historical angular direction $\phi$ using Equation (\ref{eq:phi}). This equation computes $\phi$ by summing the product of $m_{ij}$ and $A_{ij}$ for all pairs $i,j \in N$, and normalizing it by the total number of zones $|Z|$. The resulting value of $\phi$ provides an estimation of the extent to which the route's actual order deviates from the expected historical direction. We can also calculate the value of the deviation from historical heading vectors $\eta$ as defined in Equation (\ref{eq:eta}). This equation computes $\eta$ by summing the product of $w_{ij}$ and $(1 - H_{ij})$ for all pairs $i,j \in Z$, and normalizing it by the total number of zones $|N|$. Finally, we can estimate the average zone-to-zone travel time of the route $\lambda$. Equation (\ref{eq:lambda}) provides an estimation of $\lambda$ by summing the product of $m_{ij}$ and $M_{ij}$ for all pairs $i,j \in Z$, and dividing it by the maximum travel time between any two zones. The resulting value of $\lambda$ captures the overall impact of travel times on the route, considering both inter-zone travel and travel between zones and the delivery station.

\begin{equation} \label{eq:eta}
    \eta = \frac{\sum\limits_{i,j \in Z} m_{ij} \cdot (1 - H_{ij}) }{|N|}
\end{equation}

\begin{equation} \label{eq:phi}
     \phi = \frac{\sum\limits_{i,j \in N} m_{ij} \cdot A_{ij}}{|Z|}
\end{equation}

\begin{equation} 
\label{eq:lambda}
    \lambda = \frac{\sum\limits_{i,j \in Z} m_{ij} \cdot M_{ij}}{max(M_{ij})}
\end{equation}

 We can include this information within a routing optimization algorithm by modifying the objective function from Equation \ref{eq:general_objective} as follows. First, we normalize the total travel time and add it as the component normalized total time ($\tau$). Then, we add the components: historic headings deviation ($\eta$), historic angles deviation ($\phi$), and zones-to-zones travel time ($\lambda$). Each component has different magnitudes and is normalized to ensure comparable values. We assign a weighting factor $\theta_i$ to each component, reflecting its importance in the objective function. The objective function that considers travel time and information from historical routes is shown in Equation \ref{eq:historic_objective}.

 \begin{equation} \label{eq:historic_objective}
    f_o =  \theta_1 \tau + \theta_2 \eta + \theta_3 \phi + \theta_4 \lambda 
\end{equation}

We modify the single-objective CATSP solver\ref{algo:general_catsp_solver} following the Two-stage data-driven GRASP metaheuristic introduced by Mesa et al. \cite{mesa2023amazon}. The modified CATSP solver uses two phases (routines) to generate route sequences: the historical information extraction phase and the route planning phase. First, it extracts information from historical routes in the form of transition probability matrices. Then, it proceeds with the CATSP solver, using traditional routing information (travel time) and adding the transition probability matrices to plan the routes. The pseudo-code of the modified CATSP solver with historical routing information is presented in the Appendix in Algorithm \ref{algo:catsp_solver_historic_info}. For a complete explanation of the method, we refer the reader to the work of Mesa et al. \cite{mesa2023amazon}. 

 \begin{algorithm} [h]

    \KwIn{$V, T, Q$}
    
    $F$ $\leftarrow$ Create vector field ($Q$)
	
	$H_{ij}$, $A_{ij}$ $\leftarrow$ Calculate transition probability matrices ($F, V$)

    $M_{ij}$ $\leftarrow$ Calculate zone-to-zone travel-time ($V$)
    
 \caption{Extract historical information phase. Adapted from \cite{mesa2023amazon}}
    \label{algo:extract_historic_info}
\end{algorithm}

The first stage of the approach is the extract historical information phase. The \textbf{Create vector field} step constructs the vector field $F$ using historical route information from the set of historic routes $Q$. The \textbf{Calculate transition probability matrices} step leverages $F$ and its set of zones $Z$ to calculate the transition probability matrices $H_{ij}$ and $A_{ij}$ for a new route $x$ that needs planning. We condense lines 4 to 9 from Algorithm\ref{algo:general_catsp_solver} into the \textbf{Calculate zone-to-zone travel-time} method. Moving on to the second stage, the route prediction phase, the routing process of Algorithm\ref{algo:general_catsp_solver} is executed, with the addition of $H_{ij}$, $A_{ij}$, and $M_{ij}$ into the objective function $f_o$ to generate the new route sequence $x$.

In summary, we can construct a vector field by analyzing historical routes using the step vectors between consecutive zone centroids. This vector field captures the preferred directions of routes across the city, allowing us to estimate the heading or order of zones for new routes based on their proximity to the observed routes in the vector field. Finally, this information, along with stop-to-stop travel times and zone-to-zone estimated travel times, can be fed into a weighted objective function that can be used within a CASTP solver.

\subsection{Adapting the CATSP solver for solving the bi-objective routing problem}
\label{sec:adapting_solver_biobjective}

Independently of whether the method that uses visual attractiveness metrics or the one that leverages data mining information from historical routes works better, we are going to tackle the problem as a bi-objective problem. To this end, we aim to find solutions that make up the Pareto front. In this section, we show how we adapt the general framework of the CATSP solver in order to solve a bi-objective vehicle routing problem.



Let's assume we have two objective functions $f_1$ and $f_2$ for the CATSP, $f_1$ considers traveling time cost, while $f_2$ considers preferences from the drivers or decision makers. To solve this bi-objective problem, we integrate the single-objective CATSP solver from Algorithm \ref{algo:general_catsp_solver} within a heuristic box splitting algorithm \cite{matl2019leveraging} to enable the ability to generate a set of Pareto-optimal solutions (just like any multi-objective algorithm).

The heuristic box splitting (HBS) algorithm proposed by Matl et al. \cite{matl2019leveraging} is based on representing the limits of the objective space with an initial rectangle and iteratively solving and splitting the rectangle space into smaller rectangles while respecting the limits imposed on the objective function by the current rectangle. HBS can be applied to a route optimization algorithm. Therefore, we take the GRASP algorithm in Algorithm \ref{algo:general_catsp_solver}, which corresponds to lines 10 to 17, and use it as the route optimization heuristic with some modifications. The modifications consist of using the objective functions $f_1$ and $f_2$, by using $f_1$ as the objective function to minimize subject to $f_2 \leq c$ as a constraint. We present the \textbf{Route Optimization Heuristic} (ROH) in Algorithm \ref{algo:route_optimization}. 

 \begin{algorithm}[h] 
 
	\KwIn{$N, Z, L, c$}
	\nonl \hrulefill

        $cost_{best} \leftarrow \infty$

	\For{$i \leq L$}
	{   
	    Zones solution $z \leftarrow$ Random\_order($Z$)
		
		$z \leftarrow$ relocate\_zones($z, c$)
		
		Solution $x \leftarrow$ Random\_order ($z, N$)
		
		$x \leftarrow$ VND($x, neighborhoods, c$)
		
		\If{$f_1(x) < cost_{best} $ and $f_2(x) \leq c$}
		{
			$cost_{best} \leftarrow f_1(x)$
			
			$x_{best} \leftarrow x$
		}

	}
 
	\Return{$x_{best}$}
	
	\caption{Route optimization heuristic (ROH).}
    \label{algo:route_optimization}
\end{algorithm}

Now, we take ROH and the CATSP solver, and integrate them into the heuristic box algorithm. The pseudocode of the proposed bi-objective CATSP solver is presented in Algorithm \ref{algo:catsp_with_box_splitting}. The proposed algorithm uses rectangles or boxes to represent areas in the objective space that may contain non-dominated solutions. The algorithm maintains an archive of these rectangles $R_{areas}$ and a set of non-dominated solutions $A_{sol}$ that have been found. Initially, $f_2^{min}$ is a valid lower bound on objective $f_2$ (We use  $f_2^{min}=0$), $f_1^{max}$ is any upper bound on objective $f_1$ (we use $f_1^{max}=cost$(random nearest neighbor solution)), and $\delta_{min}$ is the minimal difference in objective $f_2$ when considering two feasible solutions (we use $\delta_{min}=0.01$. The other inputs of the algorithm ($V, T, L$) are the same inputs of Algorithm \ref{algo:general_catsp_solver}.

The algorithm starts by \textbf{Assigning missing zone IDs}, where the missing zone IDs of the stops in a given route are assigned. If a stop does not have a \textit{zone ID}, the algorithm assigns it the zone ID of the nearest stop (that already has a zone ID) based on the estimated travel time between them. Then, in \textbf{Calculate zone-to-zone travel-time}, the algorithm utilizes the travel time information between the stops in a given route to calculate the estimated travel time between zones $M_{ij}$. After calculating the zone centroids travel-time, the algorithm proceeds with the phase of \textbf{Routes prediction phase: Heuristic Box Splitting Two-Stage GRASP}. First, we execute the ROH of Algorithm \ref{algo:route_optimization} to attempt to minimize objective $f_1$ without assigning an upper bound to objective $f_2$ (which is the same as subjecting it to the constraint $f_2 \leq \infty$). This yields the ``upper-left” extreme point ($f_1^{min}$, $f_2^{max}$). The ``initial lower-right” extreme point corresponds to the two initial bounds ($f_1^{max}$, $f_2^{min}$). Together, these two points delimit the initial box (or rectangle). The ``true lower-right" point is progressively approximated by splitting unexamined regions of the objective space and discarding redundant regions. The proposed algorithm iteratively splits the largest remaining rectangle in half by setting the $\epsilon-constraint$ and then solves the corresponding subproblem using ROH. The boundaries of the selected rectangle determine the value of $c$, which indicates the upper bound of ($f_2 \leq c$) when solving ROH. The subproblem may be infeasible, so ROH returns NULL instead of a feasible solution. In such cases, the archive $A_{sol}$ remains unchanged, and the explored half of the rectangle is discarded, resulting in no new rectangle. If a solution $x$ is found, this resulting solution lies within the boundaries of the selected rectangle, and it will dominate a region of the rectangle. This region is removed, $R_{areas}$ is updated with the new rectangles, and $A_{sol}$ is updated with the new non-dominated solution while solutions that are dominated by $x$ are removed from it. The process is repeated until $R_{areas}$ is empty or after $n_{max}$ iterations are performed. Finally, the algorithm's output is the archive $A_{sol}$ with all the non-dominated solutions found during the execution.

 \begin{algorithm}
 
	\KwIn{$V, T, L, n_{max}$, $f_1^{max}$, $f_2^{min}$, $\delta^{min}$}
    \KwOut{Archive of non-dominated solutions $A_{sol}$}
	\nonl \hrulefill
	
    $V$ $\leftarrow$ Assign missing zone IDs ($V$)
	
	$M_{ij}$ $\leftarrow$ Calculate zone-to-zone travel-time ($V$)

    $x \leftarrow ROH(f_1,\infty)$

   $ A_{sol} \leftarrow \{x\}$ , $z^1 \leftarrow (x_1, x_2)$ , $z^2 \leftarrow (f_1^{max}, f_2^{min})$, $R_{areas} \leftarrow \{$R$(z^1, z^2)\}$

    \For{$i \leq n_{max}$}
	{   
        Find R$(y^1, y^2) \in R_{areas}$ with maximal area

      $ c \leftarrow \frac{1}{2} (y_2^1 + y_2^2)$

       $x \leftarrow ROH(f_1, c)$

       \If {$x = NULL$ or $x$ is dominated}
       {
       
        $y^2 \leftarrow (y_1^2, c)$
       
       }
        \Else {
         Update archive $A_{sol}$ with $x$

         \If{there exists \{R'$(z^1, z^2) \in R_{areas}$ $| z_1^1 < x_1 \leq z_1^2$\}}{
         
         $z* \leftarrow (x_1, max\{z_2^2, c\})$
         
         \If{ $z_2^1 - z_2^* > \delta_{min}$}
         {
         
         Add R$(z^1, z^*)$ to $R_{areas}$
         }
         
         }

        \If{there exists \{R''$(z^1, z^2) \in R_{areas}$  $| z_2^1 \geq x_2 \geq z_2^2$\}}{
        
         $z^* \leftarrow (max\{x_1, z_1^1\}, x_2$)
         
         \If{$ z_2^* - z_2^2 > \delta_{min}$}
         {
         
         Add $R(z*, z2)$ to $R_{areas}$
         
         }

         }

         Delete R' and R'' if they exist

         \ForAll {R$(z^1, z^2) \in R_{areas}$ with $z^1$ dominated by $x$}{

            Delete R from $R_{areas}$

            }
         
        }
        If {$R_{areas} = \emptyset $}{
        
        Stop
        
        }    
        
        \Return{$A_{sol}$}
        }
	\caption{Bi-objective Two-Stage GRASP with heuristic box splitting. Based on \cite{matl2019leveraging, mesa2023amazon}.}
    \label{algo:catsp_with_box_splitting}
\end{algorithm}


\begin{equation}
\label{eq:obj_1}
f_1 =  \theta_1 \tau
\end{equation}

\begin{equation}
\label{eq:obj_2}
f_2 =  \theta_2 \eta + \theta_3  \phi + \theta_4 \lambda 
\end{equation}

Later, in Section \ref{experiments}, we explain in detail the experiments and the results of applying these three approaches (Visual attractiveness metrics, Data mining from historical routes, and Bi-objective routing) to last-mile delivery planning. In summary, the single objective historical pattern mining strategy works better than the single objective visual attractiveness metrics. Therefore, we take the method for extracting historical patterns from routes and integrate it within the bi-objective algorithm, as shown in the Appendix in Algorithm \ref{algo:appendix_catsp_with_box_splitting_historical_routes}. Additionally, we construct the following objective functions that formulate the bi-objective mathematical problem. Equation (\ref{eq:obj_1}) seeks to optimize the total routing cost. Equation (\ref{eq:obj_2}) seeks to minimize the weighted differences in the routing sequence with respect to historical routes.

\section{Computational experiments}
\label{experiments}

In this section, we present and analyze the results of two sets of experiments carried out in our work. First, we compare the use of visual attractiveness metrics vs. leveraging historical information from routes to predict last-mile delivery routes. Then, taking the method with better performance in the first experiment, we enhance it to enable the solution of bi-objective problems and evaluate the algorithm's ability to approximate the Pareto front when solving the proposed bi-objective model. For both experiments, we use the set of CATSP instances provided by Amazon and MIT in the 2021 Amazon Last Mile Routing Research Challenge \cite{amazonchallenge, amazon-dataset}.

All the routing algorithms were coded in C++ and compiled using g++ 11.2.0 and Boost 1.78 library \cite{BoostLibrary}. As previously mentioned, to estimate the vector fields, the heading vectors, and the different probability matrices, we used the code provided by Mesa et al. \cite{mesa2023amazon} \footnote{Available at \url{https://github.com/mesax1/Data-driven-last-mile-routing}}, which is written in Python 3.8. All the experiments were run on an Intel Xeon Gold 6130 CPU, running at 2.10 GHz, with 64 GB of RAM, on a Rocky Linux 8.5 Operating System. All experiments were run on a single CPU core. The code repository for this work is available at \url{https://github.com/mesax1/biobjective_lastmile_routing}.

\subsection{Experiments of visual attractiveness metrics versus historical patterns mining}
\label{visualvsml}

We evaluate and compare two approaches to create routes as similar as possible to the way in which routes have been previously executed. This is with the objective of minimizing the differences between previously executed routes and newly planned routes. The two approaches are: using visual attractiveness metrics to assess the routes and data mining techniques to find patterns in the routes.

To evaluate which of the two approaches works best, we ran a set of experiments that determines which strategy generates routes that resemble the most previously executed routes. To this end, we used Amazon's dataset of last-mile delivery routes \cite{amazon-dataset} and a score similarity metric that compares two routes and outputs a scalar value that indicates their similarity.

The score similarity metric, in Equation \ref{eq:scoring_function}, calculates the similarity between two route sequences, $x$ and $B$, based on three factors: the sequence deviation $SD(x,B)$, the normalized edit distance $ERP_{norm}(x,B)$, and the number of edit operations $ERP_e(x,B)$ required to transform $x$ into $B$. If $x$ and $B$ are the same sequence, then $score(x, B)$ is 0. Additional information on how the score similarity metric is computed is given in the Appendix \ref{appendix_scoring_function}.

\begin{equation}
    \label{eq:scoring_function}
    score(x, B) = \frac{SD(x, B) \cdot ERP_{norm}(x,B)}{ERP_e(x,B)}
\end{equation}

The sequence deviation $SD(x,B)$, shown in Equation \ref{seq_dev}, measures the deviation of the position of stops in the predicted route sequence $x$ with respect to the actual route sequence $B$. Assuming both routes visit the same $n+1$ stops ($n$ stops and one depot station). First, we identify the position $g$ of each stop in sequence $x$ within sequence $B$. We then form a vector \( (g_0, g_1, ..., g_n) \). Next, we create vector $(g_0, g_1, ..., g_n)$. $SD(x,B)$ is the sum of the absolute differences between consecutive positions in this vector, divided by the total number of unique stop combinations.

\begin{equation}
\label{seq_dev}
    SD(x, B) = \frac{2}{n(n-1)}\sum_{i=1}^n(|g_i - g_{i-1}| - 1)
\end{equation}

``The number of edit operations $ERP_e(x,B)$ required to transform sequence $x$ into sequence $B$ is calculated using the \textit{Levenshtein distance}, which is a measure of the difference between two sequences introduced by Levenshtein et al. \cite{levenshtein1966binary}. Let $E = E_0, E_1, ..., E_n$ be the edit operations to transform sequence $x$ into sequence $B$. Additionally, let $C = c_0, c_1, ..., c_n$ be the normalized travel time costs associated with the operations performed. Each operation of insertion, deletion, and substitution has a unit cost."\cite{mesa2023amazon}. Then, the normalized edit distance is calculated as shown in Equation \ref{erp_norm}.

\begin{equation}
\label{erp_norm}
    ERP_{norm}(x,B) = \sum_{i=0}^n c_i E_i
\end{equation}

A lower score indicates that $x$ is more similar to $B$. A score of zero means they are exactly the same sequence. This metric was utilized in the 2021 Amazon Last Mile Routing Research Challenge \cite{amazonchallenge}\footnote{Further details of the scoring algorithm and related implementations can be found at \url{https://github.com/MIT-CAVE/rc-cli/tree/main/scoring}}.

To evaluate the performance of routing guided by historical patterns found in previous routes, we modified Algorithm \ref{algo:general_catsp_solver} following the procedure described in Section \ref{sec:adapting_solver_historical_routes}, using the same values for the parameters and the objective function weights (Equation \ref{eq:historic_objective}) as proposed by Mesa et al. \cite{mesa2023amazon}. The method consists of four weighting factors $(\theta_i)$ and two parameters that need to be calibrated based on the available data. The four components of the objective function \ref{eq:historic_objective} are assigned a weighting factor $\theta_i$. In addition, there are two parameters to consider: the length of step vectors ($\beta$) in meters for estimating heading vectors and the weighting factor ($\alpha$) for step vectors. Table \ref{table:historicparameters} outlines the values for each parameter.

\begin{table}[ht]
\centering
\begin{adjustbox}{max width=\textwidth}
\begin{threeparttable}

\caption{\label{table:historicparameters} Parameters of the data mining from historical routes method. Adapted from \cite{mesa2023amazon}.}
\rowcolors{2}{gray!25}{white}
\begin{tabular}{l l >{\bfseries}l}
\rowcolor{gray!50}
\hline
Parameter & Description & Value \\
\hline

$L$ & Number of GRASP iterations & 50  \\
$\alpha$ & Weighting factor for step vectors & 0.01  \\
$\beta$ & Length of the step vectors in meters & 200  \\
$\theta_1$ & Weighting factor of total time $\tau$ & 3  \\
$\theta_2$ & Weighting factor of historic zone headings $\eta$ & 1  \\
$\theta_3$ & Weighting factor of historic zone angles $\phi$ & 1  \\
$\theta_4$ & Weighting factor of zones time $\lambda$ & 5  \\
\hline
\end{tabular}
\end{threeparttable}
\end{adjustbox}
\end{table}

For evaluating the performance of visual attractiveness metrics, and ensuring a fair comparison, we use the same code provided by Mesa et al.\cite{mesa2023amazon} and modify it following the procedure described in Section \ref{sec:adapting_solver_visual_metrics}. We incorporate the four visual attractiveness metrics: Average Distance Compactness ($adc$), Center Distance Compactness ($cdc$), Number of Crossings ($nc$), and Bending Energy ($be$), as penalization components of the objective function. We keep the parameter $L$ with the same value. Table \ref{table:visualparameters} presents the values for each parameter.

\begin{table}[ht]
\centering
\begin{adjustbox}{max width=\textwidth}
\begin{threeparttable}

\caption{\label{table:visualparameters} Parameters of the visual attractiveness metrics method.}
\rowcolors{2}{gray!25}{white}
\begin{tabular}{l l >{\bfseries}l}
\rowcolor{gray!50}
\hline
Parameter & Description & Value \\
\hline

$L$ & Number of GRASP iterations & 50  \\
$\theta_1$ & Weighting factor of total time $\tau$ & 3  \\
$\theta_4$ & Weighting factor of zones time $\lambda$ & 5  \\
$\theta_6$ & Weighting factor of average distance compactness $\upsilon_{adc}$ & 1  \\
$\theta_7$ & Weighting factor of center distance compactness $\upsilon_{cdc}$ & 1  \\
$\theta_8$ & Weighting factor of number of crossings $\upsilon_{nc}$ & 1 \\
$\theta_9$ & Weighting factor of bending energy $\upsilon_{be}$ & 1  \\
\hline
\end{tabular}
\end{threeparttable}
\end{adjustbox}
\end{table}

We use the 3052 routes of the evaluation dataset to evaluate both methods. For the historical patterns data mining method, we execute Algorithm \ref{algo:catsp_solver_historic_info} once per route, with the parameter configuration presented in Table \ref{table:historicparameters}, considering all the components of the objective function. We then calculate the score of each predicted route, by comparing it against how the real route was executed, using the scoring metric function of Equation \ref{eq:scoring_function}. For the visual attractiveness metrics method, we execute Algorithm \ref{algo:catsp_solver_visual_metrics}, with the parameters configuration of Table \ref{table:visualparameters}. For each visual attractiveness metric, we execute the algorithm once per route, keeping the components of total time $\tau$ ($\theta_1 = 3$) and zones time $\lambda$ ($\theta_4 = 5$), and including only the single attractiveness metric that we want to evaluate with a weight of 1, e.g. the number of crossings $\upsilon_{nc}$ ($\theta_8 = 1$). Then, we calculate the score for each route using the same procedure as for the historical patterns method.

\begin{figure}[h]
  \centering
    \includegraphics[width=0.9\textwidth]{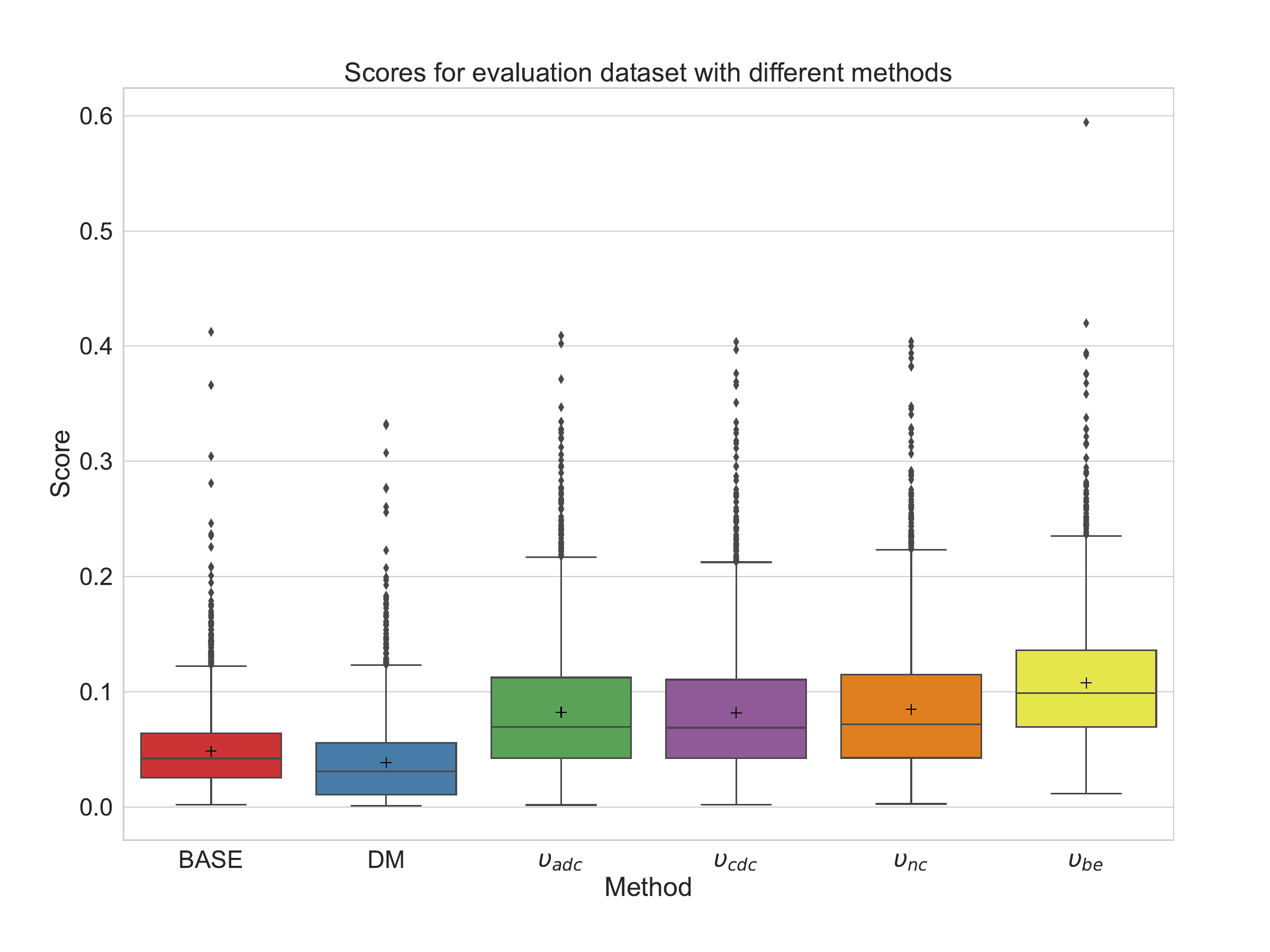}
  \caption{Distributions of the scores obtained with different routing methods on the evaluation dataset.}
  \label{fig:boxplot_eval_different_methods}
\end{figure}

We summarize the score results obtained with these experiments in Figure \ref{fig:boxplot_eval_different_methods} and Table \ref{table:results_different_methods}. The results for the historical patterns data mining method are represented with the name ``DM". The results for each visual attractiveness metric are represented by the name of the parameter, e.g. $\upsilon_{adc}$. Additionally, we include the results of only using the components of total time ($\theta_1 = 3$) and zones time ($\theta_5 = 5$) in the objective function as a baseline, represented with the name ``BASE". In Table \ref{table:results_different_methods}, for each column, we report the average route score ``Mean", the standard deviation ``Std", the best score of a route ``Min", the ``Median" score, and the worst score of a route ``Max".

\begin{table}[ht]
\centering
\begin{adjustbox}{max width=\textwidth}
\begin{threeparttable}

\caption{\label{table:results_different_methods} Results for scores obtained with different routing methods on the evaluation dataset.}
\rowcolors{2}{gray!25}{white}
\begin{tabular}{l l >{\bfseries}l l l l l}
\rowcolor{gray!50}
\hline
 &  BASE & DM   &  $\upsilon_{adc}$  &  $\upsilon_{cdc}$  &  $\upsilon_{nc}$ & $\upsilon_{be}$ \\
\hline

Mean  &                      0.0484 &            0.0413 &          0.0822 & 0.0817 &    0.0848 &    0.1076 \\
Std   &                     0.0335 &            0.0355 &           0.0544 &   0.0540 &    0.0561 &    0.0531 \\
Min   &                      0.0025 &            0.0008 &          0.0017 &     0.0020 &    0.0027 &    0.0117 \\
Median  &                    0.0428 &            0.0337 &          0.0696 &     0.0688 &    0.0716 &    0.0989 \\
Max   &                      0.3543 &            0.3249 &          0.4091 &     0.4036 &    0.4041 &    0.5942 \\

\hline
\end{tabular}
\end{threeparttable}
\end{adjustbox}
\end{table}

Visually, we can notice the performance difference between the methods. The best performance is achieved using the historical patterns data mining method, followed by the ``classical" baseline, which aims to minimize travel times, both between stops and between zones. When any single visual attractiveness metric is added to the baseline objective function, the performance worsens. The average score for the baseline is 0.0484. The ``DM" method improves the score to 0.0411, while the best average score achieved by a visual attractiveness metric is just of 0.0817 for the $\upsilon_{cdc}$, and the worst average score of 0.1076 is obtained with the $\upsilon_{be}$ metric. For this particular situation of predicting single-vehicle last-mile delivery routes, the advantage of using the historical patterns data mining method is clear. Therefore, we select this method and use it to address the bi-objective vehicle routing problem proposed in this work.

\subsection{Experiments of bi-objective vehicle routing problem}

After selecting the method for solving the single-objective VRP problem, we modified the method following the heuristic box-splitting strategy proposed by Matl et al. \cite{matl2019leveraging}. Now, instead of having the single objective function of Equation \ref{eq:historic_objective}, we have the bi-objective problem with Equation \ref{eq:obj_1} and Equation \ref{eq:obj_2}. We followed the procedure described in Section \ref{sec:adapting_solver_biobjective}, and turned Algorithm \ref{algo:catsp_solver_historic_info} into Algorithm \ref{algo:appendix_catsp_with_box_splitting_historical_routes}. Table \ref{table:biobjectiveparameters} presents the parameters of the bi-objective CATSP solver and their corresponding values.

\begin{table}[ht]
\centering
\begin{adjustbox}{max width=\textwidth}
\begin{threeparttable}

\caption{\label{table:biobjectiveparameters} Parameters of the bi-objective CATSP solver.}
\rowcolors{2}{gray!25}{white}
\begin{tabular}{l l >{\bfseries}l}
\rowcolor{gray!50}
\hline
Parameter & Description & Value \\
\hline

$L$ & Number of GRASP iterations & 50  \\
$\alpha$ & Weighting factor for step vectors & 0.01  \\
$\beta$ & Length of the step vectors in meters & 200  \\
$\theta_1$ & Weighting factor of total time $\tau$ & 1  \\
$\theta_2$ & Weighting factor of historic zone headings $\eta$ & 1  \\
$\theta_3$ & Weighting factor of historic zone angles $\phi$ & 1  \\
$\theta_4$ & Weighting factor of zones time $\lambda$ & 5  \\
$n_{max}$ & Maximum number of HBS iterations & 50 \\
$f_1^{max}$ & Initial upper bound of $f_1$  (24 hours in seconds)& 86400 \\
$f_2^{min}$ & Initial lower bound of $f_2$ & 0 \\

\hline
\end{tabular}
\end{threeparttable}
\end{adjustbox}
\end{table}

We run Algorithm \ref{algo:appendix_catsp_with_box_splitting_historical_routes} on the 3052 instances of the evaluation dataset. Due to the randomness involved in the algorithm, ten runs per instance were performed. The average running time for an instance is 45 seconds. Table \ref{table:nondominatedcount} presents the results on the average number of non-dominated solutions obtained with our bi-objective CATSP solver. The first column represents the number of non-dominated solutions found. The second column shows the average number of instances where the non-dominated solutions were found.

\begin{table}[ht]
\centering
\begin{adjustbox}{max width=\textwidth}
\begin{threeparttable}

\caption{\label{table:nondominatedcount} Number of non-dominated solutions found.}
\rowcolors{2}{gray!25}{white}
\begin{tabular}{l >{\bfseries}l}
\rowcolor{gray!50}
\hline
Number of non-dominated solutions found & Number of instances \\
\hline

1 & 604 \\
2 & 1020 \\
3 & 769 \\
4 & 397 \\
5 or more & 262 \\

\hline
\end{tabular}
\end{threeparttable}
\end{adjustbox}
\end{table}

\begin{table}[ht]
\centering
\begin{adjustbox}{max width=\textwidth}
\begin{threeparttable}
\caption{\label{table:traveltimes} Travel time differences in the Pareto front.}
\rowcolors{2}{gray!25}{white}
\begin{tabular}{l >{\bfseries}l}
\rowcolor{gray!50}
\hline
\textbf{Description} & \textbf{Value} \\
\hline
Average maximum travel time & 11812 s \\
Average minimum travel time & 11251 s \\
Average difference between maximum and minimum travel time & 602 s \\
Number of routes with more than 10 minutes maximum time difference & 1202 \\
Number of routes with more than 15 minutes maximum time difference & 569 \\
Number of routes with more than 30 minutes maximum time difference & 47 \\
Number of routes with more than 45 minutes maximum time difference & 5 \\
Number of routes with more than 60 minutes maximum time difference & 1 \\
\hline
\end{tabular}
\end{threeparttable}
\end{adjustbox}
\end{table}

From Table \ref{table:nondominatedcount}, it can be seen that the algorithm obtains at least 2 non-dominated solutions for 80.21\% (2448 instances out of 3052) of the instances. Only for 8.58\% of the instances the algorithm found 5 or more non-dominated solutions. On average, there are at least 2.60 non-dominated solutions for each instance. Three factors can explain the latter. The first one is related to how the algorithm explores the objective space: Even if the algorithm tries to minimize the objective functions $f_1$ and $f_2$, the local search of the algorithm is guided by the travel time cost, both between zones (clusters) and from customer to customer, where solutions with lower travel time cost are mainly explored. The second factor relates to how the historical routes are planned. Due to tight operations margins, last-mile delivery routes are typically planned in a way that minimizes operations costs. Therefore, information from historical routes will also trend toward minimizing travel time. Finally, as indicated by Mesa et al. \cite{mesa2023amazon}, almost 37\% of the routes are located in areas where no route in the training dataset was executed, so there is not enough information available for the objective function $f_2$ to consider.

\begin{figure}
  \centering
    \includegraphics[width=0.75\textwidth]{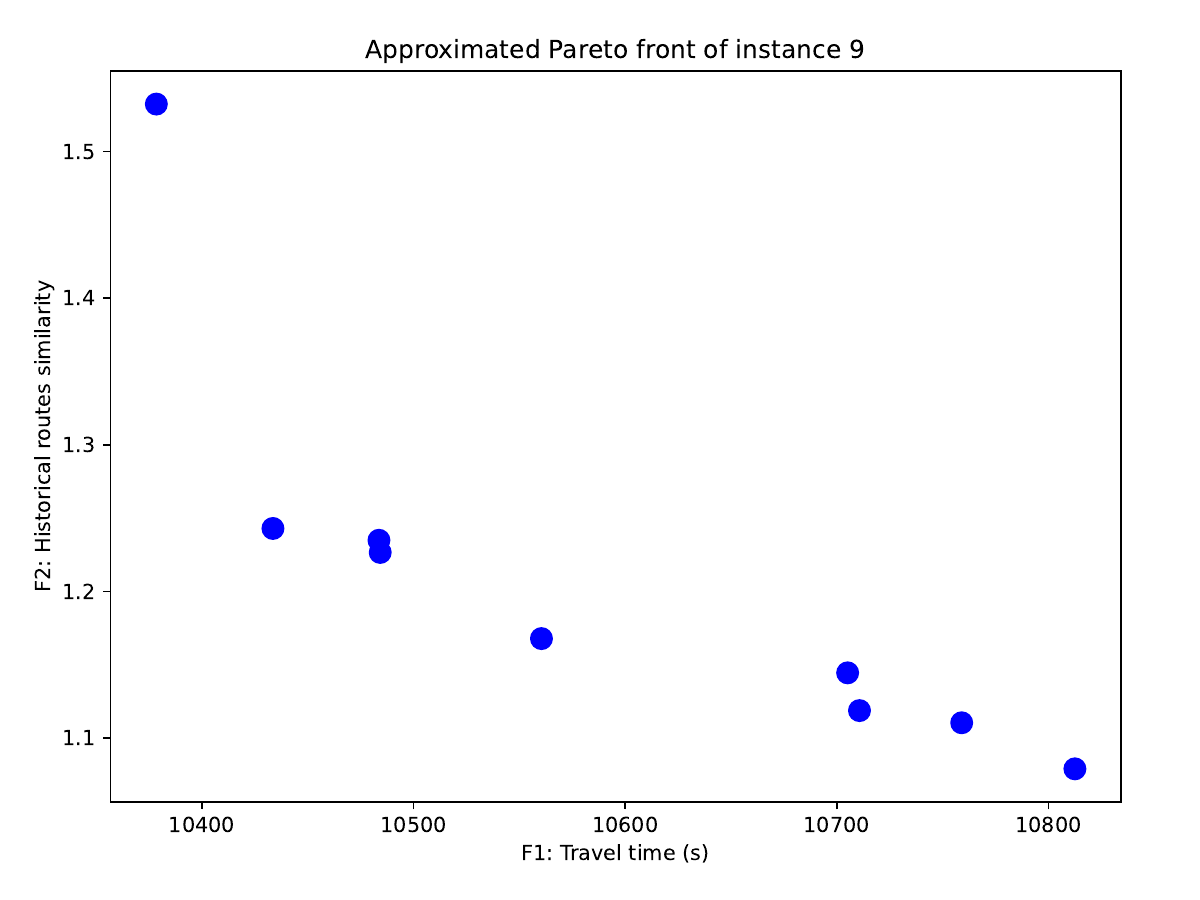}
  \caption{Approximated Pareto front for instance 9.}
  \label{fig:pareto_front_9}
\end{figure}

\begin{figure}
  \centering
    \includegraphics[width=0.75\textwidth]{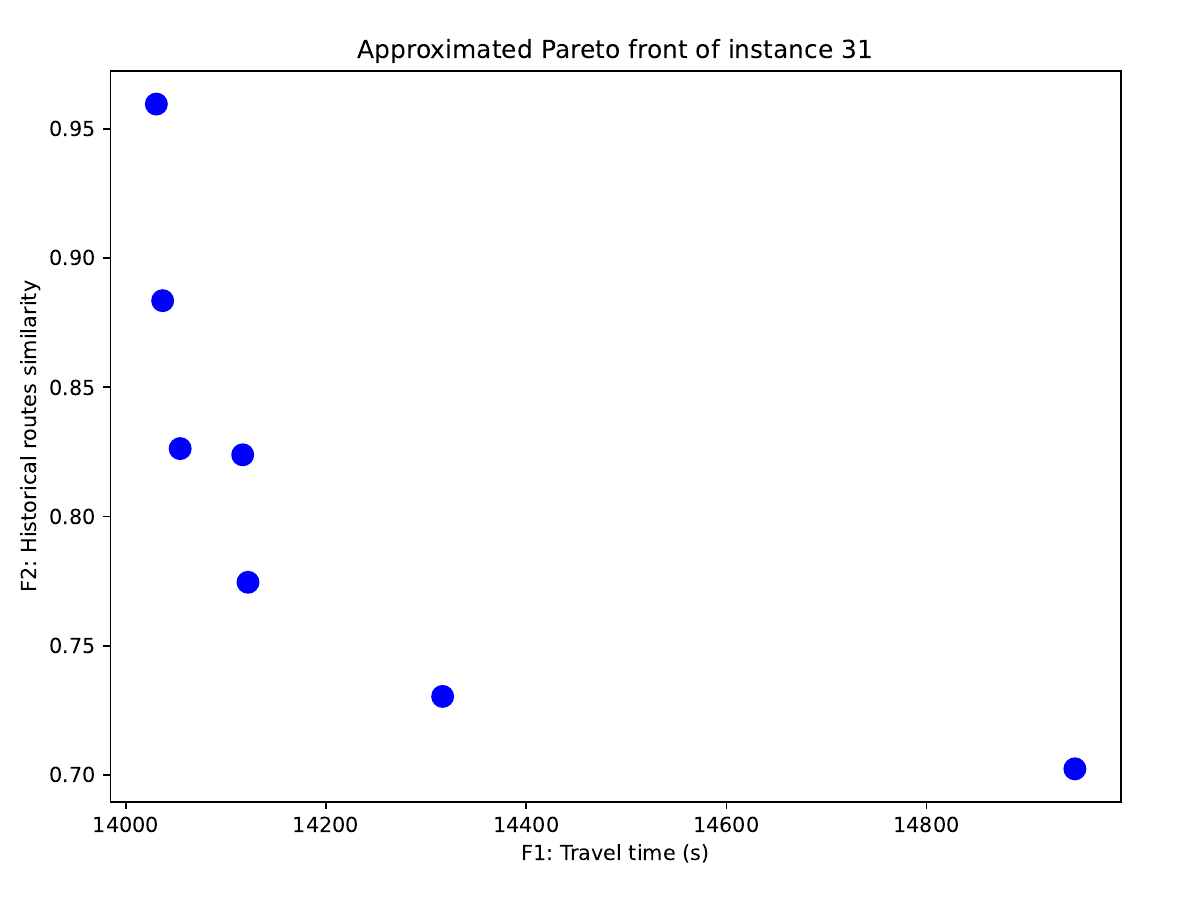}
  \caption{Approximated Pareto front for instance 31.}
  \label{fig:pareto_front_31}
\end{figure}

\begin{figure}
  \centering
    \includegraphics[width=0.75\textwidth]{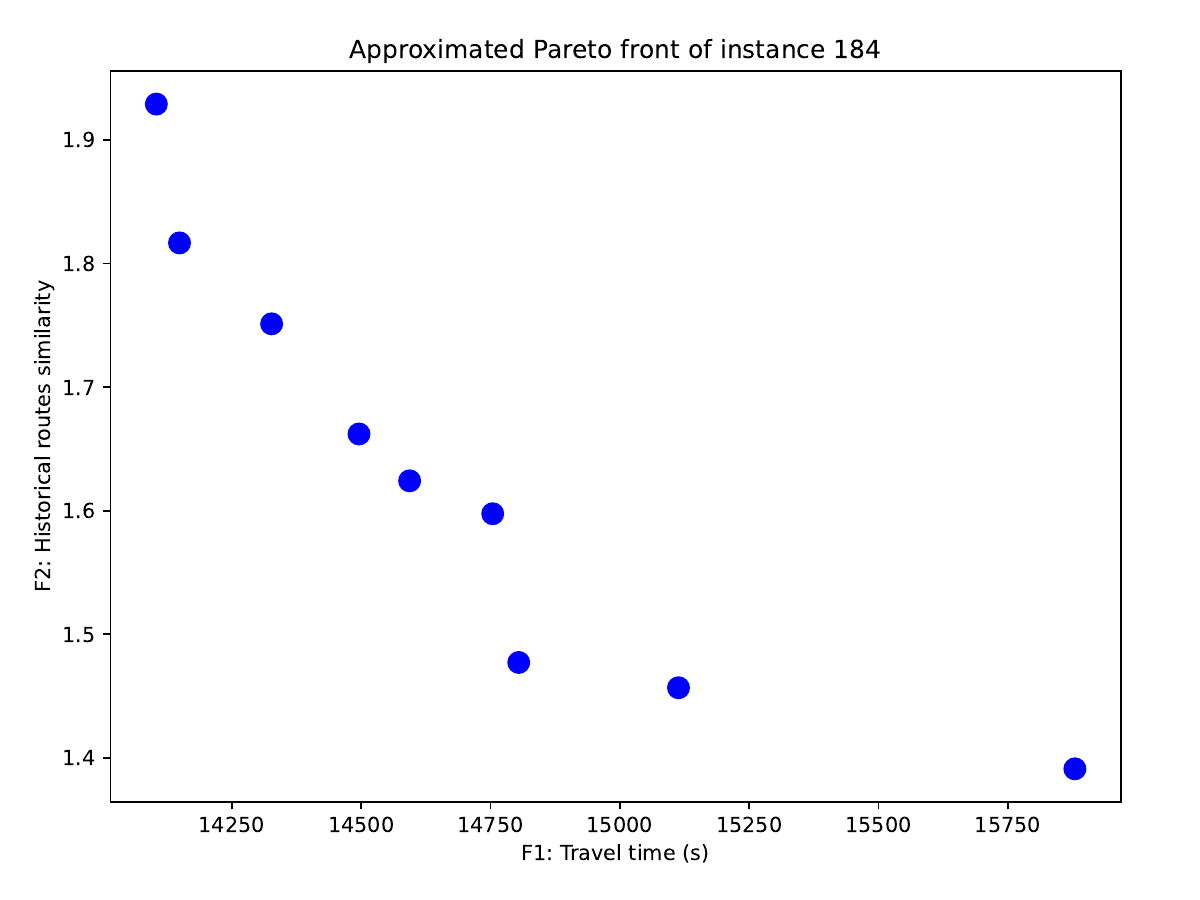}
  \caption{Approximated Pareto front for instance 184.}
  \label{fig:pareto_front_184}
\end{figure}

\begin{figure}
  \centering
    \includegraphics[width=0.75\textwidth]{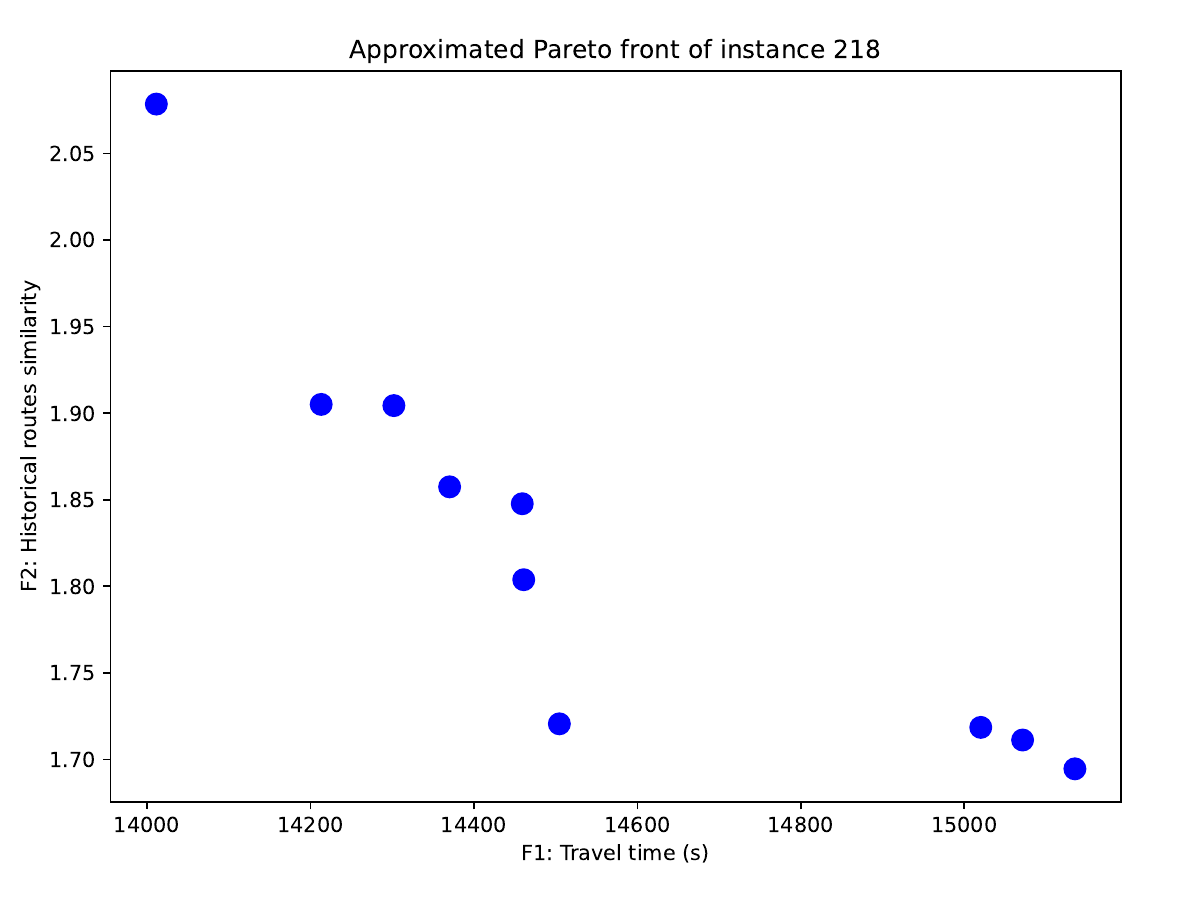}
  \caption{Approximated Pareto front for instance 218.}
  \label{fig:pareto_front_218}
\end{figure}

The approximated Pareto fronts for multiple instances are plotted in Figures \ref{fig:pareto_front_9} through \ref{fig:pareto_front_218}. The approximated Pareto fronts obtained by the proposed bi-objective CATSP solver are plotted in blue. Additionally, in Table \ref{table:traveltimes} we report information around the average travel time differences for the 2448 instances where at least two non-dominated solutions were found. When comparing the travel time cost for an instance, the average difference between the possible route found with the minimum travel time and the most similar route to the historical routes is 10 minutes. This relatively small difference can be explained by the way Algorithm \ref{algo:appendix_catsp_with_box_splitting_historical_routes} focuses on searching routes and improvements based on reducing the traveling time cost, therefore most solutions will tend to minimize travel time.

The obtained results show that the proposed algorithm is able to find non-dominated solutions. As a result, a Pareto front can be approximated. This is important because having a set of non-dominated solutions instead of a unique solution gives support to the decision-maker. For example, a decision-maker may want to maximize historical route-similarity for new drivers to promote learning familiar routes, but bearing in mind the total travel time to avoid long routes that exceed working time limits.

\section{Conclusions}
\label{conclusions}

In this work, we introduced a bi-objective vehicle routing problem with simultaneous minimization of traveling costs and differences with historical patterns. Additionally, we compared the strategies of using visual attractiveness metrics or data mining of patterns, to predict routes that resemble how routes are executed by drivers. These situations were studied using data from a real-world last-mile delivery operation. The main objective of this work is to help route planners and decision-makers to find the trade-off between travel time cost and learned patterns.
We introduce a bi-objective metaheuristic that uses a simple GRASP and a heuristic box-splitting procedure to approximate a Pareto front for each instance.

Our results when comparing visual attractiveness metrics versus data mining of historical patterns, suggest that the latter strategy is better to predict routes in the way that drivers execute them, compared to the visual attractiveness metrics found in the literature. The results of the computational experiments indicate that the algorithm is able to find only a small amount of non-dominated solutions per instance. The small number of non-dominated solutions generated per instance can be beneficial to decision-makers. Using the set of solutions provided by the Pareto front, a decision-maker can choose the routes that the drivers should follow. Selecting a route amongst two or three different options can be simpler and more time-efficient compared to selecting a route amongst 20 or 30 options. This becomes more important when the operation requires hundreds or thousands of routes that need to be dispatched daily, and the time spent deciding which route to execute for each instance also has to be minimized.

Our proposed method uses a simple GRASP as the main route optimization procedure. As a future research direction, this GRASP can be replaced by a state-of-the-art routing algorithm that can consider penalizations within its objective function. This could lead to a broader and better exploration of the solution space, which can lead to finding more solutions on the Pareto front. Another idea for further research is to consider other methods for extracting information from historical routes. For example, some other ideas presented in \cite{winkenbach2021technical} could be considered to guide the local search heuristic via constraints, as Cook et al. \cite{cook2022constrained} further explain in their work.

\section{Acknowledgments}

The authors acknowledge supercomputing resources made available by the Universidad EAFIT scientific computing center (APOLO) to conduct the research reported in this work and for its support for the computational experiments. Finally, the first author is grateful to Universidad EAFIT for the Ph.D. Grant from project 974-000005.

\section{Conflicts of Interest}
The authors declare that there are no conflicts of interest regarding the publication of this paper.


\appendix

\section{CATSP Solver with Visual Attractiveness Metrics}
\label{sec:algorithm_catsp_visual_metrics}

\begin{algorithm}
 
	\KwIn{$V, T, L, f_o$}

    $cost_{best} \leftarrow \infty$
  
	$V$ $\leftarrow$ Assign missing zone IDs to stops ($V$)

    $M = \emptyset \leftarrow $Matrix of average cluster-to-cluster (zone-to-zone) travel times
    
    \For{$g = 1, 2, \dots, m$}
	{
 
        \For{$h = 1, 2, \dots, m \quad h \neq g$}
	   {

            \For{$i \in V_g$}
	       {
 
             \For{$j \in V_h$}
	           {

                    $M_{gh} = M_{gh} + T_{ij}$
    
                    }

                }
            $M_{gh} = \frac{M_{gh}}{|V_g| + |V_h|} $
    
            }

    }

	\For{$iteration \leq L$}
	{   
	    Zones solution $z \leftarrow$ Random\_order($Z$)
		
		$z \leftarrow$ relocate\_zones($z, M$)
		
		Solution $x \leftarrow$ Randomize ($z, N$)
		
		$x \leftarrow$ VND($x, neighborhoods, T$)

        $\upsilon_k \leftarrow $ compute\_visual\_attractiveness\_metric($x$)
		
		\If{$f_o(x, \upsilon_k) < cost_{best} $}
		{
			$cost_{best} \leftarrow f_o(x, \upsilon_k)$
			
			$x_{best} \leftarrow x$
		}

	}
 
	\Return{$x_{best}$}
	
	\caption{CATSP solver with visual attractiveness metrics. Adapted from \cite{mesa2023amazon}}
    \label{algo:catsp_solver_visual_metrics}
\end{algorithm}
\clearpage
\section{CATSP Solver with Historical Routes Information}
\label{sec:algorithm_catsp_historical_routes}

\begin{algorithm}

    \KwIn{$V, T, Q, L, f_o$}
    \nonl \hrulefill
  
	$V$ $\leftarrow$ Assign missing zone IDs to stops ($V$)

    \nonl \hrulefill
    
	\nonl Extract historical information phase
	
	\nonl \hrulefill
	
	$F$ $\leftarrow$ Create vector field ($Q$)
	
	$H_{ij}$, $A_{ij}$ $\leftarrow$ Calculate transition probability matrices ($F, V$)

    $M_{ij}$ $\leftarrow$ Calculate zone-to-zone travel-time ($V$)
	
	\nonl \hrulefill

	\nonl Route prediction phase: Two-Stage GRASP

        \nonl \hrulefill

        $cost_{best} \leftarrow \infty$

	\For{$i \leq L$}
	{   
	    Zones solution $z \leftarrow$ Random\_order($Z$)
		
		$z \leftarrow$ relocate\_zones($z, M, H, A$)
		
		Solution $x \leftarrow$ Random\_order ($z, N$)
		
		$x \leftarrow$ VND($x, neighborhoods$)
		
		\If{$f_o(x, M, H, A) < cost_{best} $}
		{
			$cost_{best} \leftarrow f_o(x, M, H, A)$
			
			$x_{best} \leftarrow x$
		}

	}
 
	\Return{$x_{best}$}
	
	\caption{CATSP solver with historical routing information. Adapted from \cite{mesa2023amazon}}
    \label{algo:catsp_solver_historic_info}
\end{algorithm}

\section{Bi-objective CATSP Solver with Historical Routes Information}
\label{sec:appendix_algo_bi_objective}

\begin{algorithm}
 
	\KwIn{$Q, V, L, n_{max}$, $f_1^{max}$, $f_2^{min}$, $\delta^{min}$}
    \KwOut{Archive of non-dominated solutions $A_{sol}$}
	
    \nonl \hrulefill
	
	\nonl Extract historical information phase
	
	\nonl \hrulefill
	
	$F$ $\leftarrow$ Create vector field ($Q$)
	
	$H_{ij}$, $A_{ij}$ $\leftarrow$ Calculate transition probability matrices ($V$)
	
	\nonl \hrulefill
 
	\nonl Routes prediction phase: Heuristic Box Splitting Two-Stage GRASP

    \nonl \hrulefill

    $V$ $\leftarrow$ Assign missing zone IDs ($V$)
	
	$M_{ij}$ $\leftarrow$ Calculate zone-to-zone travel-time ($V$)

    $x \leftarrow ROH(f_1,\infty, H, A, M)$

   $ A_{sol} \leftarrow \{x\}$ , $z^1 \leftarrow (x_1, x_2)$ , $z^2 \leftarrow (f_1^{max}, f_2^{min})$, $R_{areas} \leftarrow \{$R$(z^1, z^2)\}$

    \For{$i \leq n_{max}$}
	{   
        Find R$(y^1, y^2) \in R_{areas}$ with maximal area

      $ c \leftarrow \frac{1}{2} (y_2^1 + y_2^2)$

       $x \leftarrow ROH(f_1, c, H, A, M)$

       \If {$x = NULL$ or $x$ is dominated}
       {
       
        $y^2 \leftarrow (y_1^2, c)$
       
       }
        \Else {
         Update archive $A_{sol}$ with $x$

         \If{there exists \{R'$(z^1, z^2) \in R_{areas}$ $| z_1^1 < x_1 \leq z_1^2$\}}{
         
         $z* \leftarrow (x_1, max\{z_2^2, c\})$
         
         \If{ $z_2^1 - z_2^* > \delta_{min}$}
         {
         
         Add R$(z^1, z^*)$ to $R_{areas}$
         }
         
         }

        \If{there exists \{R''$(z^1, z^2) \in R_{areas}$  $| z_2^1 \geq x_2 \geq z_2^2$\}}{
        
         $z^* \leftarrow (max\{x_1, z_1^1\}, x_2$)
         
         \If{$ z_2^* - z_2^2 > \delta_{min}$}
         {
         
         Add $R(z*, z2)$ to $R_{areas}$
         
         }

         }

         Delete R' and R'' if they exist

         \ForAll {R$(z^1, z^2) \in R_{areas}$ with $z^1$ dominated by $x$}{

            Delete R from $R_{areas}$

            }
         
        }
        \If {$R_{areas} = \emptyset $}{
        
        Stop
        
        }    
        
        \Return{$A_{sol}$}
        }
	\caption{Bi-objective Two-Stage GRASP with HBS and DM. Based on \cite{matl2019leveraging, mesa2023amazon}.}
    \label{algo:appendix_catsp_with_box_splitting_historical_routes}
\end{algorithm}

\clearpage

\section{Scoring metric}
\label{sec:scoring_metric_appendix}

The scoring metric used on the 2021 Amazon Last-Mile Routing Challenge is shown in Equation \ref{appendix_scoring_function}.

\begin{equation}
    \label{appendix_scoring_function}
    score(x, B) = \frac{SD(x, B) \cdot ERP_{norm}(x,B)}{ERP_e(x,B)}
\end{equation}

$B$ is the actual executed route sequence, and $x$ is the predicted route sequence. $D(x,B)$ measures the sequence deviation of $x$ with respect to $B$. $E_{norm}(x,B)$ denotes the edit distance weighted with a real penalty applied to sequences $x$ and $B$ using normalized travel times. $E_{op}(x,B)$ counts the number of edits required to transform sequence $x$ into sequence $B$. If both sequences are identical, $x = B$, then $score(x, B)$ is replaced with a score of 0.

The sequence deviation $D(x,B)$ is calculated as follows. Assuming both routes visit the same $n+1$ stops ($n$ stops and one depot station). First, we take each stop from $x$ and trace its position $g$ in $B$. Next, we create vector $(g_0, g_1, ..., g_n)$. Finally, we compute the sequence deviation using Equation \ref{appendix:seq_dev}.

\begin{equation}
\label{appendix:seq_dev}
    D(x, B) = \frac{2}{n(n-1)}\sum_{i=1}^n(|g_i - g_{i-1}| - 1)
\end{equation}

The number of edit operations $E_{op}(x,B)$ is calculated using the \textit{Levenshtein distance} introduced by Levenshtein et al. \cite{levenshtein1966binary}. Let $E = E_0, E_1, ..., E_n$ be the minimum number of edit operations to transform sequence $x$ into sequence $B$. Additionally, let $C = c_0, c_1, ..., c_n$ be the normalized travel time costs associated with the operations performed. Each operation of insertion, deletion, and substitution has a unit cost. Then, the normalized edit distance similarity metric is calculated as shown in Equation \ref{appendix:erp_norm}.

\begin{equation}
\label{appendix:erp_norm}
    E_{norm}(x,B) = \sum_{i=0}^n c_i E_i
\end{equation}

The more a predicted route sequence $x$ coincides with the actual route $B$, the lower the score. A score of zero means they are exactly the same sequence \footnote{Details of the scoring algorithm and related implementations can be found at \url{https://github.com/MIT-CAVE/rc-cli/tree/main/scoring}}. 

\clearpage

\end{document}